\documentclass[10pt, twocolumn]{article}

\usepackage[a4paper, top=2.5cm, bottom=2.5cm, left=1.8cm, right=1.8cm, columnsep=0.6cm]{geometry}
\usepackage[T1]{fontenc}
\usepackage[utf8]{inputenc}
\usepackage{lmodern}
\usepackage{microtype}
\usepackage{graphicx}
\usepackage{booktabs}
\usepackage{amsmath}
\usepackage[dvipsnames]{xcolor}
\usepackage[colorlinks=true, linkcolor=MidnightBlue, citecolor=MidnightBlue, urlcolor=MidnightBlue]{hyperref}
\usepackage{titlesec}
\usepackage{abstract}
\usepackage[font=small, labelfont=bf]{caption}
\usepackage{graphicx}
\usepackage{booktabs}
\usepackage{multirow}
\usepackage{makecell}
\usepackage{array}

\titleformat{\section}{\large\bfseries\scshape}{\thesection.}{0.5em}{}
\titleformat{\subsection}{\normalsize\bfseries}{\thesubsection.}{0.5em}{}

\usepackage{hyperref} 

\usepackage[backend=biber, style=apa, sorting=nyt]{biblatex}
\makeatletter
\DeclareCiteCommand{\parencite}[\mkbibparens]
  {\usebibmacro{prenote}}
  {\bibhyperref{\usebibmacro{citeindex}%
    \usebibmacro{cite}}}
  {\multicitedelim}
  {\usebibmacro{postnote}}
\makeatother

\title{\vspace{-1cm}{\large \textbf{Conversation-as-Data RAG: Comparing Graph-Based Retrieval Architectures for Cross-Meeting Question Answering}}}
\author{%
  Fernando Perez Tapia \\ \small University of Sydney \\ \small \texttt{fperezt30@gmail.com}
  \and
  Hongyi Liu \\ \small University of Sydney \\ \small \texttt{hliu2752@uni.sydney.edu.au}
  \and
  Jin Xue \\ \small University of Sydney \\ \small \texttt{jin.xue@sydney.edu.au}
}
\date{November 2026}

\begin{document}

\maketitle
\thispagestyle{empty}

\begin{abstract}
  Modern project-based organizations record what happened but rarely capture "why": the rationale behind decisions is often lost when the people involved leave. This "Contextual Blind Spot" motivates the need for retrieval systems that preserve decision-making rationale across multiple related conversations, a scenario existing RAG evaluations rarely test. This report presents CAD-RAG, a Contextual Knowledge Graph-based RAG pipeline for cross-meeting question answering, evaluated against two baselines, HippoRAG and GraphRAG, on a 32-question cross-meeting benchmark over the AMI Meeting Corpus, using both LLM-as-judge and human scoring. GraphRAG achieves the highest overall score, an advantage best explained not by model capability, but by when synthesis occurs: GraphRAG synthesizes at indexing time, while CAD-RAG and HippoRAG defer synthesis to generation, at a lower verbosity cost. All three systems handle negation-based questions well, and the LLM judge and human scoring agree on the top-ranked system (GraphRAG) but differ in score magnitude and in the ordering of the two lower-ranked systems.
\end{abstract}

\vspace{0.5em}\hrule\vspace{0.8em}

\section{Introduction}
\subsection {Motivation}

Modern project-based organizations are proficient at recording what happened (eg. decisions, milestones, deliverables),  but the why behind those decisions is rarely captured. This rationale typically lives only in the unstructured, ephemeral discourse of meetings, and is lost the moment the people involved leave the organization. We define this phenomenon as the Contextual Blind Spot: a persistent gap between documented outcomes and the reasoning that produced them. As project teams grow more distributed and turnover increases, this blind spot compounds, leaving organizations unable to reconstruct or learn from their own decision-making history. 

\begin{figure*}[!t]
  \centering
  \includegraphics[width=\textwidth]{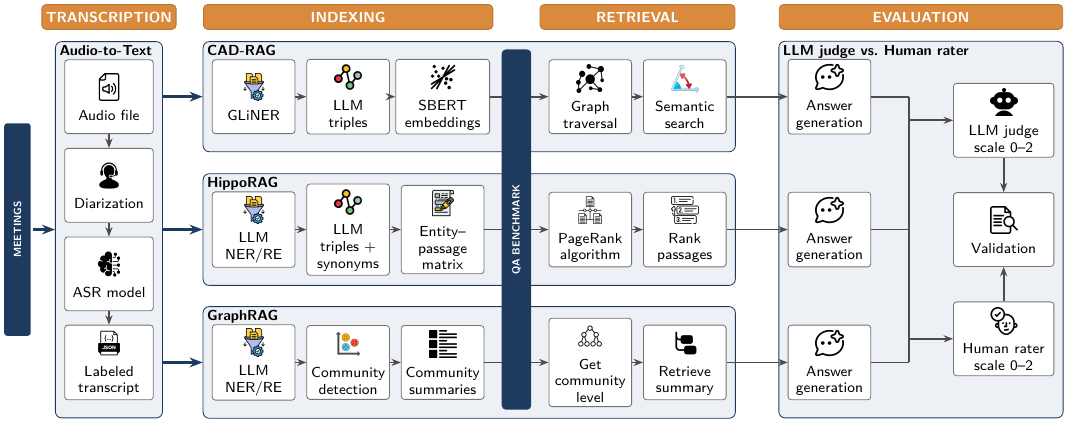}
  \caption{Overview of the experimental pipeline. The pipeline has four phases. In transcription, meeting audio is converted to a labeled transcript (diarization and ASR) shared by all three systems. In indexing, each system processes the transcript with its own method. In retrieval, each system retrieves evidence for the same 32-question benchmark and generates an answer. In evaluation, the generated answers are scored on a 0--2 scale by both an LLM judge and a human rater, and the two are compared. Icons: Flaticon.com.}
  \label{fig:pipeline}
\end{figure*}

Addressing this gap requires moving beyond static documentation toward systems that treat conversation itself as structured, queryable data and that are capable of preserving institutional reasoning as an organization's projects and personnel evolve. This report focuses specifically on the retrieval and evaluation of Contextual Knowledge Graph (CKG)-based RAG architectures as a foundational component of the systems that can preserve and query decision-making rationale in organizations. Figure~\ref{fig:pipeline} summarizes the experimental pipeline.

\subsection {Gap}

Existing Retrieval-Augmented Generation (RAG) architectures have been predominantly evaluated with open knowledge corpus datasets such as Wikipedia and news articles \parencite{retrievalaugmented_lewis_2020} to validate their effectiveness. Significant efforts have also been made to validate RAG against open-domain question answering datasets such as HotPotQA \parencite{hotpotqa_yang_2018}, 2WikiMultiHop \parencite{constructing_ho_2020}, and MuSiQue \parencite{musique_trivedi_2021}. However, less effort has been made to make meeting-domain corpus datasets available, such as AMI \parencite{ami_mccowan_2005} and the ELITR Minuting Corpus \parencite{elitr_nedoluzhko_2022}, and RAG testing over these datasets has predominantly been conducted for single-session question answering, where all relevant context resides within one coherent source \parencite{qmsum_zhong_2021}\parencite{elitrbench_thonet_2024} \parencite{getalpautomin_kang_2025}. Project-based organizational knowledge, however, is distributed across multiple related meetings over time, requiring retrieval systems to resolve references, track evolving decisions, and reconcile contradictions across meeting documents rather than within one. Whether current RAG paradigms generalize to this cross-meeting setting remains largely untested.

\subsection {Contributions}

This work makes the following contributions: (1) Conversation-as-data RAG (CAD-RAG), a contextual knowledge Graph-based RAG pipeline designed for cross-meeting question answering; (2) a curated 32-question cross-meeting QA benchmark spanning six question types with hop-distance and evidence-span annotations; (3) a comparative evaluation of CAD-RAG against two established baselines, HippoRAG and GraphRAG on the AMI Meeting Corpus. and (4) an LLM-as-judge evaluation framework for scoring multi-hop, cross-meeting answers.

\subsection {Roadmap}

The remainder of this report is structured as follows. Section 2 reviews related work across RAG architectures, knowledge graph construction, multi-hop QA benchmark design, and LLM-as-judge evaluation. Section 3 formalizes the research question and outlines the scope and limitations of this study. Section 4 describes the CAD-RAG pipeline, the construction of the cross-meeting QA benchmark, and the LLM-as-judge framework. Section 5 presents comparative results across CAD-RAG, HippoRAG, and GraphRAG. Section 6 discusses the mechanistic causes of observed performance differences, and Section 7 outlines directions for future work.

\section{Literature Review}

\subsection{RAG Architectures}

Retrieval-Augmented Generation (RAG) combines a trained language model with a retrieval step over external documents, so the model is not limited to what it learned during training and can instead ground its answers in retrieved text \parencite{retrievalaugmented_lewis_2020}. This improves performance on tasks that require substantial factual knowledge. 

Later work extended this idea in two main directions. The first focuses on making retrieval more efficient and simpler. LightRAG combines fine-grained entity-level retrieval with broader thematic context to reduce overhead \parencite{lightrag_guo_2024}. The second direction restructures retrieval around graphs instead of plain text chunks. GraphRAG builds an entity-relationship graph and groups it into hierarchical community summaries, allowing it to answer broader, summarization-style questions across large corpora \parencite{local_edge_2024}. Surveys of this graph-based RAG space organize these methods by how the graph is built, how it is retrieved from, and how retrieval connects to generation \parencite{graph_peng_2024, survey_zhang_2025}.

A separate line of work draws on how human memory works. HippoRAG models long-term memory consolidation using a personalized PageRank algorithm over a knowledge graph, allowing efficient multi-hop retrieval without needing repeated calls to the language model during retrieval \parencite{hipporag_gu_2024}. This idea has since been extended toward continual, non-parametric memory for language models \parencite{rag_gutirrez_2025}.

Direct comparisons between plain RAG and graph-based RAG report mixed results depending on the task: graph structure tends to help more on questions that require multi-hop or holistic reasoning, and less on simple fact lookup \parencite{rag_han_2025, maximizing_akar_2025}. This lines up with broader findings that language models struggle to make good use of information placed in the middle of a long retrieved context \parencite{lost_liu_2023}. 

CAD-RAG builds most directly on the HippoRAG line of work, using entity-path graph traversal as its core retrieval mechanism. GraphRAG is included in this study as a representative example of the community-summarization paradigm, rather than the entity-traversal paradigm — a distinction whose practical consequences are examined in Section 6.

\begin{table*}[ht]
\centering
\caption{Comparison of RAG Architectures Across the Pipeline}
\label{tab:rag_comparison}

\resizebox{\textwidth}{!}{%
\begin{tabular}{llllll}
\toprule
\multirow{2}{*}{\textbf{Category}} &
\multirow{2}{*}{\textbf{RAG Architecture}} &
\multicolumn{2}{c}{\textbf{Indexing}} &
\textbf{Retrieval} &
\textbf{Generation} \\
\cmidrule(lr){3-4}
 & & \textbf{Knowledge Type} & \textbf{Index Content} & \textbf{Query Method} & \\
\midrule

Basic RAG &
Basic RAG &
Plain Text &
Text Chunk &
Query Embedding &
Retrieved chunks are concatenated into the prompt \\
\midrule

\multirow{3}{*}{Graph RAG}
& GraphRAG
& Knowledge Graph
& \makecell[l]{Entity (LLM)\\
Relationship (LLM)\\
Community Detection (Alg)\\
Community Summaries (LLM)}
& \makecell[l]{Query Embedding\\
Get Community Level\\
Retrieve Summary}
& Summaries synthesized into a response and inserted into the prompt \\

\cmidrule(l){2-6}

& HippoRAG
& Knowledge Graph
& \makecell[l]{Entity (LLM)\\
Triple Extraction (LLM)\\
Synonym Linking (LLM)\\
Entity Passage Matrix}
& \makecell[l]{Get Entities in Query\\
PageRank (Alg)\\
Rank Passages}
& Top-ranked raw passages concatenated into the prompt \\

\cmidrule(l){2-6}

& CAD-RAG
& Knowledge Graph
& \makecell[l]{Entity (GLiNER+LLM)\\
Triple Extraction (LLM)\\
Embedding (SBERT)}
& \makecell[l]{Get Entities in Query\\
BFS Graph Traversal (Alg)\\
Retrieve Embeddings}
& Retrieved passages/entities concatenated into the prompt \\

\bottomrule
\end{tabular}
}
\end{table*}

\subsection{Knowledge Graph Construction}

Building a knowledge graph from unstructured conversational text depends on reliable entity and relation extraction. GLiNER treats named entity recognition as a matching problem between text spans and entity type labels, using a bidirectional transformer. This makes it flexible enough to extract entity types it was not specifically trained on, without needing task-specific fine-tuning \parencite{gliner_zaratiana_2023}. GLiNER2 extends this into a single system that can handle multiple extraction tasks at once — for example, entity extraction and classification together — guided by a schema the user defines \parencite{gliner2_zaratiana_2025}.

Alongside these structural extractors, a broader shift has taken place toward using LLMs directly for information extraction. A recent survey of this space shows that LLMs offer clear advantages in flexibility and generalization over traditional extraction pipelines: rather than requiring a separate model for each task, a single LLM can perform entity, relation, and event extraction together, guided only by natural-language task instructions \parencite{large_xu_2024}. This is conceptually close to CAD-RAG's approach, which combines GLiNER's structural extraction with LLM-based extraction guided by explicit category definitions.

The two graph-based baselines used in this study build their graphs quite differently. GraphRAG extracts entities and relationships from each text chunk using an LLM, then groups the resulting graph into communities and generates a hierarchical summary for each one, so that retrieval can happen at different levels of the community structure \parencite{local_edge_2024}. HippoRAG instead builds an open knowledge graph directly from LLM-extracted triples, without any community structure, and relies on a personalized PageRank algorithm over that graph to support multi-hop retrieval \parencite{hipporag_gu_2024}. This means the two baselines differ not only in how they retrieve, but in what kind of graph they retrieve from in the first place — a distinction that becomes relevant when comparing their construction cost and output to CAD-RAG's own graph in the Methodology section.

\subsection{Multihop QA Benchmark Design}

Benchmark design for multi-hop question answering has moved from single-document to multi-document settings over time. HotpotQA introduced multi-hop questions over Wikipedia, with supporting-fact annotations so that answers could be explained step by step \parencite{hotpotqa_yang_2018}. 2WikiMultiHop combined structured and unstructured sources to allow more thorough evaluation of the intermediate reasoning steps a model takes to reach an answer \parencite{constructing_ho_2020}. MuSiQue went further by composing multi-hop questions directly from chains of single-hop questions, which helps rule out cases where a model reaches the right answer through a shortcut rather than genuine multi-hop reasoning \parencite{musique_trivedi_2021}.

QAMPARI shifted focus to open-domain questions with many correct answers spread across multiple paragraphs, testing a different kind of reasoning — aggregation — rather than single-answer lookup \parencite{qampari_amouyal_2022}. Separately, SQuAD 2.0 introduced unanswerable questions, to test whether a system can recognize when there is not enough evidence to answer, rather than defaulting to a plausible-sounding guess \parencite{know_rajpurkar_2018}. This concern is directly relevant to this project's negation and zero-entity-retrieval question types, where the correct behavior is also to recognize absence rather than guess.

Within the meeting domain specifically, benchmark resources remain comparatively scarce. QMSum introduced query-based summarization over multi-domain meeting transcripts \parencite{qmsum_zhong_2021}, and MeetingQA proposed extractive question answering directly over meeting transcripts \parencite{meetingqa_prasad_2023}. Separate work has also looked at generating this kind of benchmark data more efficiently, to reduce the cost of building source-grounded, information-seeking dialogue datasets from meeting transcripts \parencite{efficient_golany_2024}. The AMI Meeting Corpus provides a widely used multi-party meeting dataset with rich annotation \parencite{ami_mccowan_2005}, and the ELITR Minuting Corpus offers a bilingual resource oriented toward automatic minuting \parencite{elitr_nedoluzhko_2022}. Building on ELITR, ELITR-Bench evaluates long-context language models specifically as meeting assistants \parencite{elitrbench_thonet_2024}, and a recent shared task applied RAG methods directly to question answering over meeting transcripts \parencite{getalpautomin_kang_2025}.

As noted in Section 1.2, however, evaluation in this domain has largely stayed within single-session settings, leaving cross-meeting, multi-session evaluation comparatively underexplored. This project's benchmark targets that gap directly: it draws on the question-type diversity established by \parencite{hotpotqa_yang_2018, musique_trivedi_2021, qampari_amouyal_2022, know_rajpurkar_2018}, while grounding every question in the meeting-transcript domain established by \parencite{ami_mccowan_2005, elitr_nedoluzhko_2022}.

\subsection{LLM-as-Judge Evaluation}

Using LLMs themselves as evaluators has emerged as a scalable alternative to human evaluation, particularly for open-ended questions where a single correct answer does not exist. Zheng et al. measured how often an LLM judge (GPT-4) agreed with human evaluators, and found agreement levels above 80\%, comparable to the level of agreement between two human evaluators \parencite{judging_zheng_2023}. This suggests LLM judges can serve as a reasonably reliable stand-in for human evaluation at scale.

However, the granularity of the scoring scale matters. ELITR-Bench used a 10-point GPT-4 rubric to score LLM responses, but found that scores clustered into only three distinct groups relative to gold human scores — roughly corresponding to the ranges [1,2], [3,5], and [6,10] \parencite{elitrbench_thonet_2024}. This indicates that despite the fine-grained scale offered, the evaluator could only reliably distinguish between three real levels of response quality, and the authors explicitly question the common practice of using 5-point or 10-point LLM-judge scales as a result.

This finding directly motivates this project's choice of a coarse 0/1/2 scoring scale for its LLM-as-judge framework, rather than a finer-grained rubric that would likely introduce noise without adding real distinguishing power

\section{Research Question}

\subsection{Statement}
This project asks: How well do RAG architectures perform in multi-meeting transcript scenarios? Specifically, it investigates whether existing RAG paradigms — entity-path graph traversal and community-based summarization — can effectively answer questions that require combining information across several related meetings, rather than a single document

\subsection{Sub-questions}

This central question breaks down into several more specific ones. Does entity-path traversal alone, without any summarization step, provide enough context to answer questions that span multiple meetings? Does community-based summarization lose the fine-grained detail needed to answer specific fact-lookup questions, even if it helps with broader questions? And do these architectures behave differently depending on the type of reasoning a question requires as well as the number of hops needed to reach an answer?

\subsection{Scope}

This study is scoped to the AMI Meeting Corpus  \parencite{ami_mccowan_2005} , specifically the four related meetings ES2006a through ES2006d, which form a single continuous project narrative involving the same group of speakers. All three systems under comparison (CAD-RAG, HippoRAG \parencite{hipporag_gu_2024}, and GraphRAG \parencite{local_edge_2024} ) are evaluated on the same 32-question benchmark, covering six question types with hop-distance and evidence-span annotations. Comparisons are made using an LLM-as-judge framework with a coarse 0/1/2 scoring scale and with human evaluation.

\subsection{Limitations}

Several limitations should be considered when interpreting these results. First, benchmark curation was constrained by time: while questions were constructed from the raw meeting transcripts with LLM assistance, the review process for the final question set would benefit from more thorough human revision to reduce potential bias and ensure fairness across question types. Second, the three systems were not evaluated under identical computing conditions. GraphRAG and CAD-RAG were run locally on CPU, while HippoRAG was run on Kaggle using GPU resources due to a dependency incompatibility on the local machine. This means direct comparisons of runtime and resource usage across systems should be interpreted with some caution, even though QA accuracy comparisons remain valid. Finally, the human evaluation was carried out by a single annotator, which would benefit from a more structured process (e.g. multiple raters and an agreement measure).

\section{Methodology}

\subsection{Dataset}
This study uses the AMI Meeting Corpus, specifically the four related meetings ES2006a through ES2006d \parencite{ami_mccowan_2005}. These four meetings form a single continuous project narrative involving the same four speakers, making them well suited to evaluating cross-meeting question answering, where later meetings reference decisions, rationale, or open items introduced in earlier ones. This corpus was chosen over single-session meeting datasets specifically such as ELITR Minuting Corpus \parencite{elitr_nedoluzhko_2022} because it allows questions to be created from connected information across multiple documents, rather than one.

\subsection{Pipeline Architecture}

This section describes how each system was built and configured for this study, beginning with the transcript-generation stage shared by all three systems, followed by system-specific construction and indexing details for CAD-RAG, GraphRAG, and HippoRAG respectively. Figure~\ref{fig:pipeline} (Section 1) gives an overview of the full experimental pipeline.

\textbf{Transcript Generation.} Rather than using AMI's manual transcripts directly, this project builds its own automatic speech recognition (ASR) pipeline. Using ASR rather than pre-existing manual transcripts allows the pipeline to be validated end-to-end from raw audio, which is necessary for it to generalize to real organizations, where meetings are not manually transcribed.

Speaker diarization is performed using pyannote 3.1, the current state-of-the-art open-source diarization system \parencite{bredin_2023}, using the pretrained speaker-diarization-3.1 model with a minimum segment duration filter of 0.5 seconds to remove noise segments. Transcription is performed using Whisper medium. Each diarization segment is transcribed independently, preserving speaker boundaries, with the language explicitly set to English to avoid unnecessary language-detection overhead. Because speakers do not always identify themselves by name early in a meeting, speaker labels are resolved to real names using an LLM (Llama-3.3-70B via Groq), which maps generic speaker labels to names mentioned conversationally in the transcript, falling back to generic labels when no name is mentioned.

\textbf{CAD-RAG: Knowledge Graph Construction.} Each meeting is processed individually first, producing an individual knowledge graph per meeting. Entity extraction follows the structural/semantic split introduced in Section 2.2: GLiNER extracts structurally well-defined entities (who, when, what, resource), while an LLM, guided by explicit category definitions, extracts more ambiguous semantic entities (why, what was decided) in parallel. SBERT embeddings support semantic similarity between entities and passages.

\textbf{CAD-RAG: Merge and Query Evaluation.} The four per-meeting graphs are combined in a dedicated merge step to form a single cross-meeting knowledge graph. Query evaluation is handled separately, through a dedicated entry point, allowing retrieval and generation to be iterated on without re-running the more expensive graph construction steps. Retrieval proceeds via BFS traversal over the merged knowledge graph.

\textbf{GraphRAG Configuration.} GraphRAG was indexed locally, using DeepInfra-hosted Llama-3.3-70B as the underlying LLM. Embeddings were generated using BAAI/bge-m3, hosted on the same DeepInfra endpoint, rather than the SBERT model used by CAD-RAG. This was an infrastructural rather than methodological choice: GraphRAG's embedding pipeline expects an OpenAI-compatible hosted API, whereas SBERT runs locally in-process. Integrating SBERT would have required either wrapping it in a custom API layer or modifying GraphRAG's internal embedding client. BAAI/bge-m3 was selected as the pragmatic option requiring no additional integration work.

Several of GraphRAG's default settings were kept unchanged. Other settings were changed as deliberate research decisions rather than compatibility fixes. max\_gleanings was set to 0 and entity\_types were trimmed to align with CAD-RAG's semantic categories, to support a fairer comparison between systems. concurrent\_requests was set to 4 and temperature to 0.0, to manage processing load and reduce variance across runs. Queries were run using local search entity-neighborhood retrieval, the mode most comparable to CAD-RAG's targeted graph traversal.

\textbf{HippoRAG Configuration.} HippoRAG was indexed on the full four-meeting corpus using its native pipeline. Open Information Extraction (OpenIE) is used to extract (subject, predicate, object) triples from each passage, followed by synonymy linking to merge phrase nodes that refer to the same underlying entity across different surface forms. This differs structurally from CAD-RAG's GLiNER-based extraction, which operates over a fixed, predefined entity taxonomy: HippoRAG's OpenIE step is schema-free, identifying noun phrases and relations without a constrained category set, trading extraction consistency for broader coverage. Retrieval is performed via personalized PageRank over the resulting graph, rather than the BFS traversal used by CAD-RAG.

To isolate architectural differences from confounding variables introduced by model choice, several components were matched to CAD-RAG's configuration wherever possible. OpenIE and synonymy detection used the same LLM as CAD-RAG and GraphRAG, Llama-3.3-70B-Instruct-Turbo via DeepInfra. The embedding model was set to Contriever, a roughly 110-million-parameter BERT-based encoder, in place of HippoRAG's default NV-Embed-v2 (7B); this kept the embedding backbone in a comparable weight class to CAD-RAG's SBERT, rather than introducing a large-model advantage unrelated to retrieval architecture itself.

The corpus was constructed as one passage per ASR-diarized transcript segment, prefixed with speaker identity, to preserve the same conversational context available to CAD-RAG's extractor. Indexing was performed on Kaggle rather than the local machine, due to a Windows incompatibility that prevented local indexing, requiring a two-notebook split to complete the process. The Contriever embedding model cache was stabilized as a private Kaggle dataset to avoid repeated downloads across sessions.

\subsection{QA Benchmark Construction}

A cross-meeting question answering benchmark was constructed to evaluate all three systems on a common, controlled set of questions. Questions were generated from the raw ASR-derived transcripts with LLM assistance. The final benchmark contains 32 questions spanning six question types: fact\_lookup, evolution, contradiction\_resolution, causal\_rationale, aggregation, and negation\_unresolved\_item. 

Each question was tagged with its hop distance (1-hop, 2-hop, 3-hop) which is the number of reasoning steps required to reach the answer. This allowed performance to be analyzed not only by question type but also by reasoning complexity. Each question was additionally annotated with its ground-truth evidence span, the specific passage or passages in the transcripts that support the correct answer, to support both scoring and error analysis. 

\subsection{LLM-as-Judge Framework}

Following the coarse scoring approach motivated in Section 2.4, each system's answer is scored on a 0/1/2 scale by an LLM judge, given the question, its question type and hop distance, the ground-truth answer and supporting evidence, and the system's generated answer. The judge is explicitly instructed to evaluate factual correctness and completeness relative to the ground truth, rather than writing quality or fluency.

A score of 0 (incorrect) is assigned when the answer is factually wrong, contradicts the ground truth, or confidently asserts something as resolved when the ground truth indicates it was never resolved — a distinction particularly relevant for negation\_unresolved\_item questions, where fabricating a resolution is treated as a clear failure rather than a partial success. A score of 1 (partially correct) is assigned when the answer is broadly in the right direction but is incomplete, misattributes the information to the wrong meeting, or lacks correct provenance despite the core claim being right; for aggregation questions specifically, this includes finding only some of the required items rather than all of them. A score of 2 (fully correct) requires the answer to match the ground truth's content and, where relevant, correctly identify which meeting the information came from, with aggregation questions requiring a complete list and negation\_unresolved\_item questions requiring a correct statement of absence rather than a fabricated answer.

The judge is additionally given type-specific evaluation guidance for each of the six question types, so that scoring criteria reflect what actually matters for that type of reasoning — for example, evolution questions must capture the change between a before-state and an after-state rather than a single snapshot, contradiction\_resolution questions must identify how a conflict was resolved rather than simply restating both sides, and causal\_rationale questions must capture the reason actually stated in the transcripts rather than a plausible-sounding but unsupported explanation.

The judge is prompted to return its output as a structured object containing a short justification followed by the numeric score, with the justification written first so that the scoring decision follows from an explicit comparison against the ground truth, rather than being asserted without reasoning. This is the feedback-before-verdict scoring pattern used in prior LLM-as-judge evaluation work \parencite{elitrbench_thonet_2024}.

Finally, to validate the LLM-as-a-judge scores, a human revision step was added. A human annotator, independent of the pipeline’s development, reviewed the same outputs using exactly the same evaluation criteria and rubric given to the LLM judge. The annotator saw the same inputs and scored each output on the same scale, without access to the LLM judge’s scores, so the two sets of ratings could be compared directly. This lets us check whether the automated judge’s assessments agree with human judgement and reveals where the two diverge.

\section{Results}

\subsection{Word Count per RAG system}

Answer length varied substantially across systems. Overall mean answer length was 9.03 words for HippoRAG, 84.03 words for CAD-RAG, and 264.88 words for GraphRAG. CAD-RAG's answers were roughly 9.3 times longer than HippoRAG's, and GraphRAG's roughly 29.3 times longer. Broken down by hop distance, HippoRAG and GraphRAG produced longer answers as hop distance increased, consistent with more retrieved evidence being incorporated into the answer as question complexity grows.

\begin{table}[ht]
\centering
\caption{Average Word Count by RAG system}
\label{tab:retrieval_time_hops}

\begin{tabular}{lccc}
\toprule
\textbf{\# Hops} & \textbf{Hippo} & \textbf{CAD} & \textbf{Graph} \\
\midrule
1-hop  & 3.00  & 52.13  & 240.75 \\
2-hop  & 5.18  & 100.27 & 269.82 \\
3-hop & 16.00 & 89.92  & 275.54 \\
\midrule
\textbf{Overall} & \textbf{9.03} & \textbf{84.03} & \textbf{264.88} \\
\bottomrule
\end{tabular}

\end{table}

\subsection{QA accuracy by question type}

Across the 32-question cross-meeting benchmark, GraphRAG achieved the highest mean score under both scoring methods (LLM judge: 0.75; human: 1.03). Under the LLM judge, HippoRAG (0.63) narrowly outperformed CAD-RAG (0.59); under human scoring, the performance reversed (0.66 CAD-RAG vs. 0.59 HippoRAG). Overall, these numbers may indicate these two architectures perform comparably.

\begin{table}[ht]
\centering
\caption{LLM Score of RAG by Question Type}
\label{tab:question_type_results}

\resizebox{\columnwidth}{!}{%
\begin{tabular}{lccccc}
\toprule
\textbf{Question Type} & \textbf{\# Q} & \textbf{Hippo} & \textbf{CAD} & \textbf{Graph} & \textbf{AVG} \\
\midrule
Aggregation             & 5  & 0.20   & 0.40 & 1.00 & 0.53 \\
Causal Rationale        & 6  & 0.33 & 0.33 & 0.33 & 0.33 \\
Contradiction Resolution& 4  & 0.50 & 0.25 & 0.50   & 0.42 \\
Evolution               & 5  & 1.00 & 0.40 & 0.60 & 0.67 \\
Fact Lookup             & 7  & 0.29 & 0.71 & 0.86 & 0.62 \\
Negation / Unresolved &  5  & 1.60 & 1.40 & 1.20 & 1.40 \\
\midrule
\textbf{Overall} & \textbf{32} & \textbf{0.63} & \textbf{0.59} & \textbf{0.75} & \textbf{0.66} \\
\bottomrule
\end{tabular}
}
\end{table}

\begin{table}[ht]
\centering
\caption{Human Score of RAG by Question Type}
\label{tab:human_question_type_results}

\resizebox{\columnwidth}{!}{%
\begin{tabular}{lccccc}
\toprule
\textbf{Question Type} & \textbf{\# Q} & \textbf{Hippo} & \textbf{CAD} & \textbf{Graph} & \textbf{AVG} \\
\midrule
Aggregation               & 5  & 0.20  & 0.40 & 1.20 & 0.60 \\
Causal Rationale          & 6  & 0.17 & 0.33 & 1.00 & 0.50  \\
Contradiction Resolution  & 4  & 0.50 & 0.50 & 0.50  & 0.50 \\
Evolution                 & 5  & 1.00 & 0.40 & 0.80 & 0.73 \\
Fact Lookup               & 7  & 0.29 & 0.71 & 1.00 & 0.67 \\
Negation / Unresolved     & 5  & 1.60 & 1.60 & 1.60 & 1.60 \\
\midrule
\textbf{Overall} & \textbf{32} & \textbf{0.59} & \textbf{0.66} & \textbf{1.03} & \textbf{0.76} \\
\bottomrule
\end{tabular}
}
\end{table}

Broken down by question type, GraphRAG's lead was most pronounced on aggregation (1.00 vs. CAD-RAG 0.40, HippoRAG 0.20 by LLM judge, with a similar pattern under human scoring) and fact\_lookup (0.86 vs. CAD-RAG 0.71, HippoRAG 0.29 by LLM judge, with a similar pattern under human scoring). Evolution achieved the second-highest overall mean score of any question type across the benchmark, behind only negation\_unresolved\_item. All three systems scored highest on negation\_unresolved\_item (n = 5) of any question type under the LLM judge (CAD-RAG 1.40, HippoRAG 1.60, GraphRAG 1.20); under human scoring, all three systems scored identically at 1.60

\section{Discussion}

\subsection{Pre-Synthesis timing might explain GraphRAG's lead}

GraphRAG's overall lead is best explained by a difference in when synthesis occurs in the pipeline, rather than by any difference in underlying model capability: all three systems generate answers using the same base LLM (Llama-3.3-70B-Instruct-Turbo). CAD-RAG and HippoRAG both retrieve comparatively raw material (subgraph passages and entity/phrase-node paths, respectively) and defer synthesis to generation time, leaving the generator to assemble a coherent answer from fragmentary evidence. GraphRAG's community summarization approach \parencite{local_edge_2024} performs a substantial part of the synthesis work at indexing time, meaning it is often surfacing an already-coherent narrative rather than constructing one from scratch at query time. This advantage is clearest on aggregation (GraphRAG 1.00 vs. CAD-RAG 0.40, HippoRAG 0.20 by LLM judge; a similar pattern under human scoring) and fact\_lookup (0.86 vs. 0.71 vs. 0.29) — question types that reward completeness across multiple mentions rather than single-fact precision.

\subsection{The cost of pre-synthesis is verbosity}

This advantage comes with a measurable cost in answer length. Overall mean answer length was 9.03 words for HippoRAG, 84.03 words for CAD-RAG, and 264.88 words for GraphRAG. CAD-RAG's answers were roughly 9.3 times longer than HippoRAG's, and GraphRAG's roughly 29.3 times longer. HippoRAG and GraphRAG produced longer answers as hop distance increased, consistent with more retrieved evidence being synthesized into the answer as question complexity grows. HippoRAG's low word count should not be read as a sign of concise, high-quality answers in isolation: a substantial portion of its short outputs are null or near-null ("None") responses arising from retrieval failure rather than deliberate brevity.

\subsection{Corpus structure explains strength on evolution}

Evolution achieved the second-highest overall mean score of any question type across the benchmark, behind only negation\_unresolved\_item. This is more plausibly explained by the structure of the AMI Meeting Corpus itself than by any one system's design. The ES2006a–d meetings follow an inherently iterative, sequential design process — functional design, conceptual design, and detailed design are each revisited across four separate meetings, with earlier proposals explicitly revised or replaced later on. Evolution is therefore not a rare or subtle discourse pattern that a system must work hard to detect — it is close to the dominant narrative shape of the corpus, with before/after states typically stated as clear, separable facts at different timestamps rather than requiring inference. This favorable data structure likely raises the ceiling for this question type independent of retrieval architecture

\subsection{Negation handling is a shared strength in all three systems}

All three systems scored highest on negation\_unresolved\_item (n = 5) of any question type under the LLM judge (CAD-RAG 1.40, HippoRAG 1.60, GraphRAG 1.20). Under human scoring, this result becomes even more striking: all three systems score identically at 1.60. This convergence under human judgment supports the interpretation that all three architectures are relatively capable of recognizing and correctly reporting the absence of a resolution rather than fabricating one, and that this is a genuine shared strength rather than an artifact of any single system's tendency toward short or null answers.

\subsection{The LLM Judge is reliable for identifying the top system, not for close comparisons}

At the aggregate level, the LLM judge and the human rater agree that GraphRAG performs best (LLM 0.75 vs. human 1.03). They differ in two respects. First, the human rater scored GraphRAG substantially higher than the LLM judge did (a gap of 0.28 on the 0–2 scale), most visibly on causal\_rationale (0.33 vs. 1.00). Second, they order the other two systems differently (LLM: HippoRAG 0.63 > CAD-RAG 0.59; human: CAD-RAG 0.66 > HippoRAG 0.59). The differences between these two systems are small (0.04 and 0.07) and, with 32 questions, should not be interpreted as a meaningful ranking under either method. The LLM judge therefore appears reliable for identifying the clearly strongest system at this study's scale, but its scores should be treated with caution when systems are close.

\section{Future Work}

\textbf{Testing whether synthesis timing alone drives the performance gap}. This study's central finding is that GraphRAG's advantage stems from when synthesis happens, not model capability (Section 6.1). Evaluating an approach like LazyGraphRAG, which defers summarization to query time rather than performing it upfront, would help isolate how much of this gap is attributable to synthesis timing alone.

\textbf{Scaling GraphRAG to a larger, more varied corpus.} GraphRAG was indexed locally on CPU under configuration constraints (Section 4.2), across only four related meetings. Testing whether its community-summarization advantage holds on larger or more varied meeting corpora would clarify how well this finding generalizes.

\textbf{Strengthening judge reliability.} The LLM judge and the human rater agreed on the top-ranked system but diverged in score magnitude (notably for GraphRAG) and in the ordering of the two lower-ranked systems. A larger human-verification study, with multiple annotators and a formal agreement measure, focused specifically on close comparisons between systems, would clarify how much weight to place on small score differences.

\textbf{Improving benchmark rigor}. A manually-sourced question subset, verified independently of any system's own extraction pipeline, would further reduce the risk of benchmark bias noted in Section 3.4.

\textbf{Validating transcript quality formally.} Transcript quality was assessed qualitatively rather than via word error rate (WER) against AMI's manual transcripts. A formal WER baseline, paired with entity-level extraction accuracy comparison, would isolate how much of CAD-RAG's performance is attributable to the ASR pipeline itself.

\textbf{Extending toward the broader VRI vision.} This report scoped its contribution to CKG-based RAG retrieval (Section 1.3). The broader vision — RLHF-based decision support and a human-in-the-loop framework — remains open, with a lightweight agent demo wrapping CAD-RAG as a natural first step.

\printbibliography[title={References}]

\end{document}